\def\year{2021}\relax
\documentclass[letterpaper]{article} 
\usepackage{aaai21}  
\usepackage{times}  
\usepackage{helvet} 
\usepackage{courier}  
\usepackage[hyphens]{url}  
\usepackage{graphicx} 
\usepackage{natbib}  
\usepackage{caption} 
\usepackage[american]{babel}

\usepackage[utf8]{inputenc} 
\usepackage{amsfonts}       
\usepackage{amssymb}
\usepackage{amsmath,amsthm}
\usepackage{array,multirow}
\usepackage{nicefrac}       
\usepackage{microtype}      
\usepackage{microtype}
\usepackage{graphicx}
\usepackage{subfigure}
\usepackage{booktabs} 

\RequirePackage{algorithm}
\RequirePackage{algorithmic}

\newtheorem*{theorem*}{Theorem}

\usepackage[normalem]{ulem}

\newcommand{\eat}[1]{}

\newcommand{\parfactor}{\texttt{parfactor}}

\title{Relational Neural Markov Random Fields}
\author{
    Yuqiao Chen , 
    Sriraam Natarajan , 
    Nicholas Ruozzi
    \\
}
\affiliations{
    University of Texas at Dallas\\
    800 W Campbell Rd, Richardson, Texas, USA\\
    \{yuqiao.chen, sriraam.natarajan, nicholas.ruozzi\}@utdallas.edu

}

\begin{document}

\maketitle

\begin{abstract}
Statistical Relational Learning (SRL) models have attracted significant attention due to their ability to model complex data while handling uncertainty. However, most of these models have been limited to discrete domains due to their limited potential functions. We introduce Relational Neural Markov Random Fields (RN-MRFs) which allow for handling of complex relational hybrid domains. The key advantage of our model is that it makes minimal data distributional assumptions and can seamlessly allow for human knowledge through potentials or relational rules. We propose a maximum pseudolikelihood estimation based learning algorithm with importance sampling for training the neural potential parameters. Our empirical evaluations across diverse domains such as image processing and relational object mapping, clearly demonstrate its effectiveness against non-neural counterparts.


\end{abstract}

\section{Introduction}
Statistical relational learning (SRL) models \citep{getoor2007,raedt2016} have gained popularity for their ability to learn and reason in the presence of noisy, uncertain, rich and structured relational data. While expressive, these models face a significant challenge when learning from real data. One such challenge arises when learning from hybrid data: for predicate/relational logic based methods, significant feature engineering is necessary to make learning both effective and efficient, and most learning methods discretize the data \citep{boostedRDN,boostedMLN} or make restrictive assumptions on the learned structure \citep{hybridRDN}.

In classical propositional (feature vector based) domains, approaches that combine the benefit of neural and graphical models have recently attracted attention. While specific network structure or training procedures differ, these methods generally employ conditional random fields (CRF) that are parameterized by neural nets, e.g., in computer vision \citep{liu2015, knobelreiter2017}. Neuro-Symbolic learning has become quite popular where neural networks are combined with a specific logic-based formalism \citep{garcez2020,raedt2020,qu2019} that combine logical reasoning and neural learning.

Here, we adopt an alternative approach that embeds neural networks directly into (relational) graphical models. Specifically, we consider the parametric factor (parfactor) graphs formalism \citep{poole2003,Braz2005} and employ neural potential functions that are flexible enough to handle hybrid (relational) data. In contrast to many existing approaches, our resulting model, dubbed relational neural Markov random fields (RN-MRFs), does not make any restrictive assumptions on the data distribution, e.g., multivariate Gaussian assumptions, nor is it dependent on a specific neural network architecture. Instead the expressivity of the procedure can be appropriately controlled by tuning the network structure/activation functions. 


We make the following key contributions. (1) We introduce a general combination of traditional SRL models and neural nets by introducing neural potential functions inside parfactor graphs. RN-MRF does not make any distributional assumptions and can handle propositional, semi-relational and relational domains. (2) We present ways for seamless integration of rich human knowledge in the form of informative potentials and/or relational rules. (3) We present an efficient training procedure for RN-MRFs using maximum pseudolikelihood estimation. (4) Finally, we demonstrate the efficacy of RN-MRFs against non-neural potential model baselines on several real domains. 

The rest of the paper is organized as follows. After covering the background, we present our RN-MRF model and outline our approach for incorporating human knowledge. We then describe the learning procedure, which is then evaluated on several real domains.

\section{Related Work}

The combination of graphical models and neural networks has been extensively explored in  image processing and related areas. The work of \citet{pmlr-v9-do10a} and \citet{xiong2020} on combining neural networks and Markov random fields are the most relevant to the approach herein. \citet{pmlr-v9-do10a} propose neural potential functions in the context of conditional random fields (CRF), where the neural network learns a feature representation, given the conditional values, of a log-linear model. Model weights are then chosen via maximum likelihood estimation.  This approach was specifically designed for discrete random variables and additional care is needed in the continuous case to ensure normalizability of the model as well as to ensure that the resulting model fitting procedure can be implemented efficiently. A similar approach was also explored in the relational setting by \citet{abs-1905-13462} where they model the possible world distribution with neural networks.

\citet{xiong2020} consider learning MRF models with neural net potentials of the form $\exp(nn(x))$, which allows both continuous and discrete domains. Their approach uses the Bethe free energy to perform approximate variational inference for use in the computation of the MLE gradient. Their approach can be applied to learn general MRFs, but can be quite slow in relational domains. As the approximate inference procedure can be unstable, the method also requires an iterative averaging procedure to ensure convergence.

Alternative approaches that combine MRFs/CRFs and neural networks have used neural networks to select the parameters of a CRF. These approaches have been particularly popular in computer vision applications in which large-scale models are common, e.g., \citep{liu2015, zheng2015, knobelreiter2017}. In these settings, the CRF are often set to have a simple structure and restrictive potential functions in order to make (approximate) inference practical. Still other approaches have demonstrated that there is a close connection between approximate MAP inference and recurrent neural networks \citep{WuLT16}.

\section{Preliminaries}
A Markov random field (MRF) consists of a hypergraph $G=(\mathcal{V}, \mathcal{C})$, with variable nodes $\mathcal{V}$ and a set of hyperedges $\mathcal{C}$. Each node $i \in \mathcal{V}$ is associated with a variable $x_i$ with domain $\mathcal{X}_i$. We consider hybrid MRFs in which $\mathcal{X}_i$ could be either discrete or continuous. Each hyperedge $c \in \mathcal{C}$ is associated with a non-negative potential function $\phi_c: \mathcal{X}_c \rightarrow \mathbb{R}_{\geq 0}$, where $\mathcal{X}_c = \cup_{i \in c} \{\mathcal{X}_i\}$ is the union of variables in the hyperedge $c$. Often, the nonnegative potential functions are represented in exponential form as $\phi_c(x_c) = \exp f_c(x_c)$, with $f_c: \mathcal{X}_c \rightarrow \mathbb{R}$. An MRF defines a joint probability distribution over joint variables $x \in \cup_{i \in \mathcal{V}} \{\mathcal{X}_i\}$
{\small
\begin{align*}
p(x) = \frac{1}{{Z}} \prod_{c \in \mathcal{C}} \phi_c(x_c)
= \frac{1}{{Z}} \exp \left(\sum_{c \in \mathcal{C}} f_c(x_c)\right),
\end{align*}
}
where the normalization term $Z = \sum_x \prod_{c \in \mathcal{C}} \phi_c(x_c)$ (summation is replaced by integration for continuous variables). 



\textbf{Fitting MRFs to Data:} Often, the potential functions are restricted to specific functional forms determined by a fixed set of parameters, $\theta$. Given $M$ training data points, $x^{(1)},\ldots,x^{(M)}$, the MRF and the corresponding distribution $p(x)$ can be fit to data via maximum likelihood estimation (MLE) by applying gradient ascent to maximize likelihood,
{\small
$$
l(x^{(1)},\ldots,x^{(M)}; \theta) = \prod_{m=1}^M p(x^{(m)}; \theta).
$$
}
The computation of the gradient in each step requires exact or approximate inference over the whole model. Exact inference is intractable in all but the simplest of models, and in large graphical models, e.g., relational models with a large number of instances, even approximate inference procedures can be computationally intensive, especially in the case of hybrid models. 

Alternatives to MLE, such as the pseudo likelihood (PL) ~\citep{besag74}, try to avoid the expensive inference step.  Specifically, the PL method approximates the joint distribution as a product of univariate conditional distributions.
{\small
$$
p(x; \theta) \approx \prod_{i \in \mathcal{V}} p(x_i \mid x_{\mathcal{V} \setminus i})
= \prod_{i \in \mathcal{V}} p(x_i \mid MB_i),
$$
}
\vspace{-1em}
where
{\small
$$
p(x_i \mid MB_i)
= \frac{\exp \sum_{c \supset i} f_c(x_i \mid x_{c \setminus i})}
{Z_i(MB_i)},
$$
}
 $MB_i = \cup_{c \supset i}\{x_c \setminus x_i\}$ is the Markov blanket of node $i$, and $Z_i$ is the partition function, which ensures that $p(x_i \mid MB_i)$ is a valid conditional probability distribution. To find the model parameters that maximize the PL, we apply the PL approximation on the loglikelihood,
{\small
\begin{align*}
\log l(M; \theta) &\approx \frac{1}{M} \sum_m \sum_{i \in \mathcal{V}} \log p(x_i^{(m)} \mid MB_i^{(m)}; \theta).
\end{align*}
}
Computing the gradient with respect to the parameters yields 
{\small
\begin{align} \label{eq:pmle-gradient}
\nabla \log l(M; \theta)
&= \frac{1}{M} \sum_m \sum_{i \in \mathcal{V}}
\sum_{c \supset i} \Big[\nabla f_c(x_c^{(m)}; \theta_c) \nonumber \\
&- \mathop{\mathbb{E}}_{p(x_i \mid MB_i^{(m)}; \theta)}
\left(
\nabla f_c(x_i, x_{c \setminus i}^{(m)}; \theta_c)
\right)\Big].
\end{align}
}
The first term of the gradient of the log-pseudolikelihood involves the gradient of the log-potentials with respect to $\theta_c$. The second term, which comes from the gradient of $\log Z_i$,  involves an expectation with respect to the conditional distributions in the likelihood approximation. 
In the discrete case, the expectation is usually computed by enumerating all possible assignments.  In the continuous case, the integral may not be computable in closed form and may need to be estimated numerically. 

\textbf{Relational MRFs:} Our key goal is to model the relations between attributes of objects by MRFs, which we represent as {\em relational Markov random fields}. We take an approach similar to lifted first-order models \citep{poole2003, choi2010}. 
In our relational MRFs, we use $atom$s to refer the attributes of objects compactly. For example, an $atom$ could be $smoke(P)$ or $friend(P, P')$, indicating that a person smokes or that a person has friendship with another person, where $P$ and $P'$ are both logical variables that could be instantiated as any possible person. Given a set of logical variables, $L$, a substitution $\delta$ is an assignment of instances to $L$. Take the $smoke(P)$ predicate as an example, if the substitution $\delta = \{P \to John\}$, the instantiated atom is $smoke(John)$.

Similar groups of attributes and/or objects are represented in the form of parametric factors $\parfactor(\phi, A, L, C)$, where $\phi$ is the potential function, $A$ a set of $atom$s, $L$ a set of logical variables, and $C$ is the constraint deciding which substitutions could be allowed. A relational MRF is defined by $\mathcal{F}$ a set of $\parfactor$s, and is equivalent to MRF after grounding (instantiating atoms with all possible substitutions). Similar to MRFs, a relational MRF defines a joint distribution over $RV(\mathcal{F})$ the set of all variables in the grounded graph.
{\small
$$
P(RV(\mathcal{F})) = \frac{1}{{Z}} 
\prod_{h \in \mathcal{F}} \prod_{\delta \in \Delta_{h}} \phi_{h}(A_{h}\delta),
$$
}
where $\Delta_h$ is the set of all possible substitutions to logical variables $L$ of parametric factor $h$, and $A_h\delta$ is the set of grounded variables obtained by grounding $A_h$ with $\delta$.

\section{Relational Neural Markov Random Fields}
We now introduce relational neural Markov random fields (RN-MRFs), a generic MRF model that combines the expressiveness of neural networks and the representational power of relational MRFs. For RN-MRFs, we propose to use neural networks to model the log-potential functions. This type of potential function can be arbitrarily expressive, hopefully allowing us to capture complex relationships among the model variables.  The expressiveness can be tuned by altering the neural network structure, the activation functions, and the training technique, e.g., using dropout.  
We introduce two types of neural potential functions. First, for continuous domains with explicit boundaries, say $[-1, 1]$ and $[0, 1.5]$, the potential function is defined as \vspace{-0.5em}
$$
\phi_{nn}(x_c) = \exp{nn(x_c)}
$$
where $nn(x_c)$ is an artificial neural network, e.g., MLP, with input dimension equal to dimension of the clique $c$ that takes each variable $x_i, i \in c$ as input to the neural net. For typical choices of activation functions, e.g., ReLU, the resulting potential function will be integrable over the domain of interest.
For unbounded domains, such as $(-\infty, +\infty)$, the potential function might not be normalizable, i.e., the corresponding integral may not exist. As a simple example, if $nn(x)$ is a linear function of $x$, the potential function is not integrable over $(-\infty, \infty)$ and it can attain arbitrarily large values depending on the choice of $x$. As a result, this potential function does not correspond to a probability distribution, which creates representational issues as well as algorithmic issues. To ensure normalizability, we propose to have potential functions of the following form.\vspace{-0.75em}
$$\phi(x_c) = \phi_{0}(x_c) \cdot \phi_{nn}(x_c),$$ 
where $\phi_{0}(x_c)$ is a carefully chosen helper distribution that can be used to ensure normalizability. As an example, if the neural net consists only of ReLU activations, then the resulting neural network will be piecewise linear. Thus, the product of a Gaussian helper potential and the neural net potential will be piecewise-Gaussian (and hence normalizable as the sum of a negative quadratic function with a linear term is still a negative quadratic function). This is the approach to normalizability that we will adopt. If, in addition, the helper distribution is easy to sample from, we will see that it can be used to estimate the integrals that are required to compute the gradient of the log-pseudolikelihood.

Similar to relational MRFs, RN-MRFs are defined by a tuple $parfactor(\phi_{0}, \phi_f, A, L, C)$, where $\phi_f$ is not limited to neural network potentials and can include other types of potentials such as Markov Logic Network (MLN) rules \citep{richardson2006}. Bounded domains do not require helper functions, and they can be set to uniform distributions over the appropriate domains. The joint distribution corresponding to an RN-MRF can be factorized as a product of a joint neural potential and a helper distribution. \vspace{-0.5em}
{\small
\begin{align*}
P(RV(\mathcal{F})) &= \frac{1}{{Z}} 
\prod_{h \in \mathcal{F}} \prod_{\delta \in \Delta_{h}} \phi_{nn}(A_{h}\delta) \cdot \phi_{0}(A_{h}\delta) \\
&= P_{nn}(RV(\mathcal{F})) \cdot P_{0}(RV(\mathcal{F}))
\end{align*}
\vspace{-2em}
}
\subsection{Encoding Human Knowledge}
One of the key attractive features of RN-MRFs is the ability to encode human knowledge in three different ways:
\begin{enumerate}
  \item Altering the structure of the model, i.e., assuming certain dependencies among a set of object attributes by creating a $parfactor$ among a set of $atom$s.
  \item Designing the potential functions, i.e., by specifying the helper function, mapping features, or tuning the structure of the neural network.
  \item Adding (weighted) logic rules that define new relationships between features.
\end{enumerate}

Consider a simple image denoising task in which the goal is to predict the original image given the noisy image. Below, we describe a possible RN-MRF to model relationships among pixels in this scenario.  We use the notation \textit{helper}:\textit{potential} to describe the model.
{\small
\begin{align} \label{rn-mrf:denoising}
LG(1, 0, 1) &: \phi_{nn1}(|val(P1) - val(P2)|) : nb(P1, P2)\nonumber\\
LG(1, 0, 1) &: \phi_{nn2}(|obs(P) - val(P)|)
\end{align}
}
where the first (parfactor) rule states that there exists a dependency between the values of adjacent pixels ($\{P, P1, P2\}$), and the second rule states that there is a relationship between observed and true values of every pixel $P$. 

For the helper distributions, we typically use either categorical distributions for discrete domains, multivariate Gaussians (or mixtures of Gaussians) for continuous domains, and categorical Gaussians for hybrid domains. In the  example, the helper function is specified as a two dimensional linear Gaussian (LG)  with $slope=1$, $intercept=0$, and $variance=1$, which provides a proposal distribution. The parameters of the helper functions can also be learned from the data if desired.

The above rules provide an example of the first mode of human knowledge integration. These two rules also include the second mode of  knowledge integration where the relationships are precisely defined. For instance, the first rule specifies that the helper distribution is a linear Gaussian. The second rule also defines a linear Gaussian helper distribution on the absolute difference between the observed and true values of every pixel. Typically, when dealing with two pixels $P1$ and $P2$, they will both be employed as two different inputs for a neural net. However, the first parfactor allows for a preprocessing step that computes the absolute difference between the pixels and uses that as the input. This is akin to feature mapping for standard supervised learning.

The third method to encode the human knowledge is to specify $parfactor$s with (soft) logical rules that define potential functions with a tunable weight parameter. The higher the weight, the higher the probability of the rule being true for a given data set \citep{richardson2006}. For instance, consider a robot mapping domain where the task is to predict the type of an observed segment $S$ given the length and depth of all segments. Length and depth are continuous features, while type is discrete $\langle$'W', 'D', 'O'$\rangle$, representing Wall, Door, and Other. We can express this in the RN-MRF formalism by introducing the following potentials/helpers.

{\small
\vspace{-1em}
\begin{align*}
U &: \phi_{nn}(length(S), depth(S), type(S)) \\
U &: \phi_{mln1}(length(S)>0.5 \Rightarrow type(S)\text{!='W'}) \\
U &: \phi_{mln2}(type(S1)\text{='D'} \Rightarrow type(S2)\text{!='D'}) : nb(S1, S2)
\end{align*}
\vspace{-1em}
}

where $U$ stands for uniform distribution, to denote that the no helper function is used.  

The first $parfactor$ uses the neural network potential to model the conditional type distribution given the segment's length and depth. Since this dataset is small, using only the data observations to train the neural nets could result in overfitting. The second $parfactor$ specifies that if the length of segment is larger than $0.5$, it should be of type $Wall$. The logical formula can be computed given the length and the type, resulting in a binary value True/False (or 0/1), indicating whether or not the logical formula is satisfied. 

The potentials are of the form $\phi_{mln}(x_c) = \exp[w \cdot logic(x_c)]$, where $w$ is a learnable parameter represents the strength of the rule. If the rule is satisfied, the potential value will be $\exp{w}$, otherwise it is $1$. The potential could also be used in representing the interrelationship between objects. For example in the third $parfactor$, the rule means that for a pair of adjacent segments, if one is of type $Door$, the other one should not be of the same type. This captures the knowledge that it is less likely for two doors be next to each other.

\section{Learning RN-MRFs}
While MLE could be used to learn the model parameters, computing the gradient of the log likelihood requires inference, which is intractable for large MRFs or models with complicated potential functions. In the relational setting, the size of the grounded RN-MRF is typically too large to admit efficient inference, and we are interested in models in which the potential functions are effectively arbitrary. Therefore, we propose to learn our model with maximum pseudolikelihood estimation instead. The MPLE approach learns the model parameters by maximizing the log-pseudolikelihood given the data, usually with gradient ascent. In the computation of the gradient of the log-pseudolikelihood, as in \eqref{eq:pmle-gradient}, two terms are considered: the gradient of the log-potential functions and the expectation of the log-potentials with respect to the local conditional distributions. For training the parameters of the neural networks, we can first calculate the gradient of the pseudolikelihood with respect to the output of the neural net and then apply back propagation to obtain the gradient of the network parameters.

While computing the first piece of the gradient is straightforward, calculating the expectation term is nontrivial as there is not a closed form equation for computing the expectation of an arbitrary function with respect to an arbitrary continuous distribution. We propose to use (self-normalized) importance sampling for the calculation.
\vspace{-0.5em}
{\small
\begin{align} \label{eq:expectation-approximation}
\mathop{\mathbb{E}}&_{p(x_i \mid MB_i^{(m)})}
\nabla nn_c(x_i , x_{c \setminus i}^{(m)}) \nonumber \\
&= \int_{-\infty}^{\infty} p(x_i | MB_i^{(m)}) \cdot \nabla nn_c(x_i, x_{c \setminus i}^{(m)}) dx_i \nonumber \\
&\approx \frac{1}{\sum_{n \sim Q} \frac{b_i(x_i^{(n)})}{Q(x_i^{(n)})}}
\sum_{n \sim Q} \frac{b_i(x_i^{(n)})}{Q(x_i^{(n)})} \cdot \nabla nn_c(x_i^{(n)},x_{c \setminus i}^{(m)}),
\end{align}
}
where $b_i(x_i) = \prod_{h \supset i} \phi_h(x_i, x_{h \setminus i}^{(m)})$, and $Q$ is a proposal distribution such that $Q(x_i) >0$ whenever $b_i(x_i) >0$. In the case that the variable $i$ has bounded domain, the proposal distribution could be chosen to be a uniform distribution over the domain. 

In case of unbounded domains and/or non-trivial helper functions, the proposal could be the product of helper functions that make up the conditional distribution, i.e., $Q(x_i) = \prod_{h \supset i} \phi_{0_h}(x_i, x_{h \setminus i}^{(m)})$. As the potentials in RN-MRFs are the product of helper functions and the neural potentials, the computation of the ratio $\frac{b_i(x_i)}{Q(x_i)}$ can be simplified.
{\small
\begin{align*}
\frac{b_i(x_i)}{Q(x_i)} &= \frac{\prod_{h \supset i} \phi_{0_h}(x_i, x_{h \setminus i}^{(m)}) \cdot \phi_{nn_h}(x_i, x_{h \setminus i}^{(m)})}{\prod_{h \supset i} \phi_{0_h}(x_i, x_{h \setminus i}^{(m)})} \\
&= \prod_{h \supset i} \phi_{nn_h}(x_i, x_{h \setminus i}^{(m)})
\end{align*}
}
The expectation can then be approximated as
{\small
\begin{align*}
\frac{\sum_{n \sim Q} \prod_{h \supset i} \phi_{nn_h}(x_i, x_{h \setminus i}^{(m)}) \cdot \nabla nn_c(x_i^{(n)}, x_{c \setminus i}^{(m)})}
{\sum_{n \sim Q} \prod_{h \supset i} \phi_{nn_h}(x_i, x_{h \setminus i}^{(m)})}.
\end{align*}
}

A na\"ive method to train the graphical model would be to compute the gradient of the parameters for each local variable distribution, sum them to build the full gradient, and perform one step of gradient ascent. However, this approach would require many passes of neural net forward/backward operations. In an RN-MRF, many of the factors share the same potential function, making parallelization possible.
To increase efficiency, we consider creating an aggregated input matrix $\{x_c^{(m)}\}_{data} \cup \{x_i^{(n)} \times x_{c \setminus i}^{(m)}\}_{samples}$ for each neural network potential, which includes both the data points and sampling points used in the approximation of the gradient. Then, we could run one pass of feed forward operation and compute the corresponding gradient for each data point.
Finally, we back propagate the gradient through the neural net and update the network parameters with gradient ascent.

Additionally, given the expressive power of neural networks, modeling distributions using neural networks can be prone to overfitting, especially on continuous domains where the learned distribution could have high peaks at the data points and be zero outside of them. To reduce the chance of severe overfitting, popular methods such as weight decay and drop out could be applied. However, we propose to modify the neural networks by  clamping the output layer, i.e., $clamp(nn(x), a, b)$ bounds the output from below by $a$ and above by $b$, to prevent the learned potentials from taking values that are too extreme. In back propagation, nonzero gradients are not propagated if $nn(x) > b$ or $nn(x) < a$.

A complete description of the RN-MRF learning algorithm can be found in Algorithm~\ref{alg:learn-rn-mrf}. During each iteration, substitutions of atoms are sampled and the Markov blanket of the resulting grounded variables are also grounded through unification. After that, the aggregated data is computed, followed by a forward pass of the neural nets. The gradient of the log pseudo likelihood w.r.t. the output of the neural net is computed and backpropagate through the whole network. Finally, parameters are updated using standard gradient ascent.

\begin{algorithm}[t]
   \caption{Learning RN-MRF}
   \label{alg:learn-rn-mrf}
\begin{algorithmic}[1]
   \STATE {\bfseries Input:} A RN-MRF $\mathcal{F}$, set of trainable potentials $\Phi$, and dataset $\mathcal{M}$
   \STATE {\bfseries Return:} Set of learned potential functions
   
   \REPEAT
       \STATE Uniformly draw a subset of substitution $\Delta_s$ for each atom $A$ and obtain the grounded variable set $\mathcal{V}_s$
       \STATE Find the unified substitutions $\Delta_u$ and get the grounding $\mathcal{V}_u$
       \STATE Uniformly draw a batch of data frames $\mathcal{M}_s \subset \mathcal{M}$
       
       \STATE Initialize the aggregated input matrix $\mathcal{D}_c$ for $\phi_c \in \Phi$
       
       \FOR{each data frame $m \in \mathcal{M}_s$}
           \FOR{each variable $i \in \mathcal{V}_s$}
               \STATE Let proposal $Q(x_i) = \prod_{h \supset i} \phi_{0_h}(x_i,x_{h \setminus i}^{(m)}))$
               \STATE Sample N number of $x_i^{(n)} \sim Q$
               \FOR{each $\phi_c$ where $c \supset i$}
                   \STATE Add sample $x_c^{(m)}$ to $\mathcal{D}_c$
                   \STATE Add samples $x_i^{(n)} \times x_{c \setminus i}^{(m)}$ to $\mathcal{D}_c$
               \ENDFOR
           \ENDFOR
       \ENDFOR
       \STATE Run forward pass for each $\phi_c \in \Phi$ with data $\mathcal{D}_c$
       \STATE Compute the gradient w.r.t the neural network output
       \STATE Run back propagation for each $\phi_c \in \Phi$
       \STATE Update each $\theta_c$ with gradient ascent
   \UNTIL{Convergence}
\end{algorithmic}
\end{algorithm}

\section{Experiments}
We now present empirical evidence of  the flexibility of RN-MRFs by considering different tasks/data domains. We aim to answer the following questions explicitly.
{\small
\begin{description}
\item[Q1:] Do RN-MRFs provide an effective data representation in continuous or hybrid domains?
\item[Q2:] How well do the neural potentials model complicated (high dimensional and multi-modal) dependencies?
\item[Q3:] Can human knowledge be easily incorporated to improve the learned RN-MRF models?
\end{description}
}
We selected several different problem domains: image denoising with grid structured model, modeling the joint distribution of Iris dataset, and relational object mapping with human knowledge. For each experiment, we use different baselines according to the domain, including Gaussian and/or Categorical Gaussian models, Neural CRF \citep{pmlr-v9-do10a}, and expert/hand-created models, such as hybrid MLNs \citep{wang08}. Notice that we do not compare against carefully engineered method such as Convolutional Neural Networks that work on specific tasks as the goal is too show the generalized effectiveness of RN-MRF across variety of domains.

We report the $\ell_1$ and $\ell_2$ error of MAP predictions for continuous domains and the MAP prediction accuracy rate and F1 score for discrete domains. For continuous and hybrid domain inference, we use expectation particle belief propagation (EPBP) \citep{lienart15}, which use sample points for approximating the continuous BP messages with a dynamic Gaussian proposal. 
All algorithms were run on a single core of a machine with a 2.2 GHz Intel Core i7-8750H CPU and 16 GB of memory, and were implemented in Python3.6 and source code is available on Relational-Neural-Markov-Random-Fields github repository.

\begin{figure}[t]
    \centering
    \includegraphics[width=1.0\columnwidth]{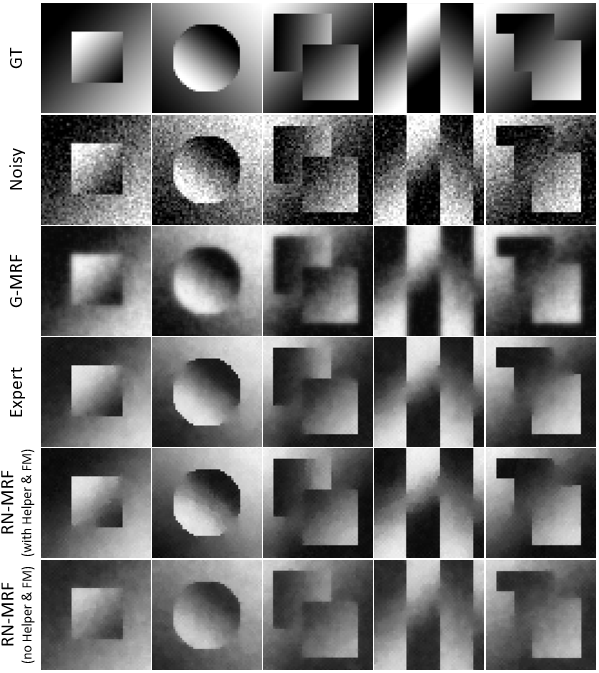}
    \caption{\footnotesize Sample outputs of different models on the simple image denoising task. G-MRF stands for Gaussian MRF. FM stands for feature mapping.}
    \label{fig:image_denoising}
    \vspace{-1em}
\end{figure}

\subsection{Image Denoising}
In this experiment, we showcase the application of our RN-MRF model in image domain with a simple image denoising task (an image has been corrupted with Gaussian noise), which requires the model to learn the relationship between neighboring pixels and observations. The reason that we choose this domain is two fold: (1) variables have continuous domains, and (2) the optimal potential functions cannot be easily represented with a simple distribution, e.g., Gaussian. For this task, the dataset was created by adding zero-mean, Gaussian noise with variance of 0.3 to 90, 300 x 400 images and 6, 50 x 50 synthetic images. 

\begin{table}[t]
     \caption{\footnotesize Comparison of different models on image denoising task.}
\centering
\scalebox{0.8}{
\begin{tabular}{|c|c|c|}
\hline
\multicolumn{1}{|c|}{Model} & \multicolumn{1}{c|}{$\ell_1$ Error} & \multicolumn{1}{c|}{$\ell_2$ Error} \\
\hline
RN-MRF(with Helper \& FM) & $89.987$ & $5.318$ \\
RN-MRF(with only FM) & $94.342$ & $5.961$ \\
RN-MRF(no Helper \& FM) & $246.531$ & $32.429$ \\
Gaussian-MRF & $125.255$ & $14.369$ \\
Expert model & $100.331$ & $6.962$ \\
\hline
\end{tabular}}
\label{tab:image_denoising}
\vspace{-1em}
\end{table}

We consider a grid structured MRF/CRF, with one layer of random variables for the input noisy image and one layer of random variables modeling the original image. We model two types of dependencies, a dependency that connects the noisy observation with the corresponding random variable in the original image and a pairwise dependency between two random variables that correspond to neighboring pixels in the original image. To represent these dependencies, we define an RN-MRF as in \eqref{rn-mrf:denoising}. Both the potentials are modeled with linear Gaussian helper function and with the feature mapping $|x_1 - x_2|$. Both neural net potentials $\phi_{nn1}$ and $\phi_{nn2}$ have two hidden layers, with 64 neurons for the first layer and 32 neurons for the second layer.

To investigate the modeling power of RN-MRF, we compare our model with Gaussian-MRFs, which have potential functions of the form $\exp(.5(x - \mu)\Sigma(x - \mu)^T)$ where the mean vector $\mu$ and covariance matrix $\Sigma$ are learnable parameters, and a hand-crafted model \citep{lienart15}:
{\small
\begin{align*}
\phi_{unary} &= \mathcal{N}(x_{obs} - x_{val}; \theta_{unary}) \\
\phi_{pairwise} &= 
\begin{cases}
  \exp(-\alpha \cdot |x_1 - x_2|) & |x_1 - x_2| < \beta \\
  \exp(-\alpha \cdot \beta) & |x_1 - x_2| \geq \beta
\end{cases}
\end{align*}
}
where the unary potential assumes the noise of the observation follows a Gaussian distribution and the pairwise potential encourages two neighboring pixels to have similar value when their observations are close. Additionally, we investigate the effectiveness of helper functions and feature mapping by evaluating the RN-MRF model both with and without a linear Gaussian helper function and without both the feature mapping and the helper function.

All models were trained using PMLE with the Adam optimizer for 5000 training iterations (until convergence). For each iteration, the algorithm randomly samples 100 variables for computing the aggregated input matrix, and approximates the expectation term with 20 sampling points. For inference, we used EPBP with a Gaussian proposal distribution and approximate the messages with 50 sampling points. We report the average image $\ell_1$ and $\ell_2$ error.

Figure~\ref{fig:image_denoising} visualizes the performance of the different approaches on sample test data points.  Gaussian MRF produces blurry results as the multivariate Gaussian function fails to fully capture the dependency between adjacent pixels. With neural potential functions, our RN-MRF model was able to perform similarly to the expert specified model. Similar conclusions can be obtained by inspecting the $\ell_1$ and $\ell_2$ errors on the test set, see Table~\ref{tab:image_denoising}.  This suggests that the RN-MRF framework is effective at modeling relationships between the pixels (\textbf{Q1}).

To assess the effectiveness of helper functions, observe that RN-MRF (with Helper \& FM) produces a more accurate result as compared to RN-MRF with only feature mapping. This is likely due to the benefits of the helper distribution when estimating the gradients. Similarly, in the RN-MRF model with neither helper distributions nor feature mapping, the accuracy degrades significantly. This suggests that the domain knowledge introduced by this feature mapping, in this case requires the potential functions to be symmetric, can have a substantial impact on the learned model. In Figure~\ref{fig:image_denoising}, the RN-MRF without FM appears to overfit the training data as it produces dimmer images (most pixels in the training images are gray). These observations suggest that adding simple domain observations can greatly improve the performance of the PMLE approach (\textbf{Q3}). 
\begin{table*}[!ht]
     \caption{Learning multi-dimensional hybrid potentials for the UCI Iris dataset.}
\centering
\scalebox{0.8}{
\begin{tabular}{|l|c|c|c|c|c|}
\hline
\multicolumn{1}{|c|}{Model} & \multicolumn{1}{c|}{Class (Accuracy)$\uparrow$} & \multicolumn{1}{c|}{Petal-Width $(\ell _2)\downarrow$} & \multicolumn{1}{c|}{Petal-Length $(\ell _2)\downarrow$} & \multicolumn{1}{c|}{Sepal-Width $(\ell _2)\downarrow$} & \multicolumn{1}{c|}{Sepal-Length $(\ell _2)\downarrow$} \\
\hline
Helper \& Neural Potential & $0.97 \pm 4.44e-5$ & $0.029 \pm 3.46e-6$ & $0.054 \pm 1.37e-5$ & $0.057 \pm 1.01e-5$ & $0.082 \pm 1.51e-5$\\
Neural Potential & $0.98 \pm 1.56e-4$ & $0.032 \pm 1.67e-5$ & $0.085 \pm 6.7e-5$ & $0.113 \pm 3.06e-5$ & $0.114 \pm 5.91e-5$\\
CG Potential & $0.933 4.72e-4$ & $0.061 \pm 4.17e-5$ & $0.136 \pm 4.54e-4$ & $0.186 \pm 1.02e-3$ & $0.717 \pm 2.16e-2$\\
FC-NN & $0.953 \pm 2.04e-3$ & $0.031 \pm 7.24e-5$ & $0.058 \pm 5.11e-5$ & $0.078 \pm 5.78e-5$ & $0.108 \pm 3.29e-4$\\

\hline
\end{tabular}}
\label{tab:single_potential}
\vspace{-1em}
\end{table*}

\subsection{Effectiveness of neural potentials}
In the simple denoising task, hand-crafted potentials are already well explored, but for more complicated domains, designing suitable potential functions would be difficult, even for domain experts. In RN-MRFs, we could utilize the expressiveness of neural potential function and effortlessly define the dependency among variables. In this experiment, we setup a single potential domain, where the graphical model has only one factor that connects all variables. It directly models the joint distribution of all the features of an object. For comparison, we design a categorical Gaussian potential function, which assigns weight $w_d$ to each possible assignment of discrete variables, and models the distribution of continuous variables with a multivariate Gaussian $\mathcal{N}_d$,  conditioned on the assignment to the discrete variables, $d$. The output of the potential function is equal to $w_d \cdot \mathcal{N}_d(x_c)$, where $\sum_{d} w_d = 1$. 

To further explore the effects of helper functions, we include a comparison of neural potentials with/without a categorical Gaussian helper function whose parameters are learned from the data. As a baseline, we also compare RN-MRFs with a standard neural network models that takes one of the features as a target and the other features as input. The aim of including this baseline is simply to show that performance does not degrade significantly in the RN-MRF framework. 

For this comparison, we use the UCI Iris data set. We use PMLE with 30 sampling points for training (3000 iterations) and EPBP (20 sampling points) for inference. A separate standard neural net is trained for each possible target feature. Both the neural potentials and the standard neural nets have 2 hidden layers with 64 and 32 neurons, respectively. All reported results are the average performance under 5-fold cross validation. The results of these experiments can be found in Table~\ref{tab:single_potential}.
The Categorical Gaussian model is less accurate when predicting both discrete and continuous values than any of the neural models. The similar performance of the RN-MRF models versus the standard neural nets suggests that the approximations introduced by the training and inference procedures do not degrade the expressiveness of the model. Recall that a different neural network is trained for each prediction task, while a single RN-MRF is fit (\textbf{Q1}).  The RN-MRF model with the helper function yields noticeably lower $\ell_2$ error in the prediction of continuous features. Again, this is likely because the helper provides a better sampling proposal as compared to the uniform distribution.

\begin{table}[t]
     \caption{\footnotesize Comparison of different models on robot mapping task.}
\centering
\scalebox{0.8}{
\begin{tabular}{|c|c|c|c|c|}
\hline
\multicolumn{1}{|c|}{Model} & \multicolumn{1}{c|}{Accuracy} & \multicolumn{1}{c|}{F1(Wall)} & \multicolumn{1}{c|}{F1(Door)} & \multicolumn{1}{c|}{F1(Other)} \\
\hline
RN-MRF & $0.876$ & $0.944$ & $0.821$ & $0.809$ \\
RN-MRF(no MLN) & $0.852$ & $0.912$ & $0.807$ & $0.781$ \\
RN-CRF & $0.901$ & $0.945$ & $0.88$ & $0.836$ \\
RN-CRF(no MLN) & $0.88$ & $0.935$ & $0.842$ & $0.812$ \\
Neural-CRF & $0.884$ & $0.961$ & $0.842$ & $0.782$ \\
CG-MRF & $0.762$ & $0.838$ & $0.703$ & $0.667$ \\
Expert HMLN & $0.761$ & $0.815$ & $0.807$ & $0.487$ \\
\hline
\end{tabular}}
\label{tab:robot_mapping}
\vspace{-1em}
\end{table}

\subsection{Robot Mapping}
To investigate the power of RN-MRF in modeling hybrid relational data and the usefulness of using a weighted logic model to capture human knowledge, we consider a real-world relational robotic map building domain \citep{limketkai2005}, where the goal is to build a labeled object map of indoor spaces from a set of laser range-scanned line segments defined by their endpoint coordinates. In this experiment, each line segment can be mapped to one of three different types of objects \{'Wall', 'Door', 'Other'\}. Similar to \citet{wang08}, we preprocess the endpoint data by computing the segments' length, depth, and angle, which are considered as evidence. We also include the $neighbor$ predicate used by \citet{wang08}: $neighbor(s_1, s_2)$ is true if the minimum endpoint distance between segment $s_1$ and $s_2$ is under a specified threshold. 

We chose this domain for several reasons: (1) it is relational and has both continuous and discrete features, (2) some relationships between features can be encoded in the form of logical language while some relationships are more complicated and can be represented with neural potentials and (3) the training data size is small -- adding human knowledge/domain expertise should help to prevent overfitting. To predict the segment type, we define the following RN-MRF.
{\small
\begin{align*}
U &: \phi_{nn1}(length(S), depth(S), angle(S), type(S)) \\
U &: \phi_{nn2}(depth(S_1)-depth(S_2), type(S_1), type(S_2)) \\
U &: \phi_{mln1}(angle(S)>30^\circ \Rightarrow type(S)\text{!='W'}) \\
U &: \phi_{mln2}(angle(S)>89^\circ \Rightarrow type(S)\text{='O'}) \\
U &: \phi_{mln3}(type(S1)\text{='D'} \Rightarrow type(S2)\text{!='D'}) : nb(S1, S2)
\end{align*}
\vspace{-1em}
}

Neural potential $nn1$ models the local relationship among the length, depth, angle, and type, which in principle would be sufficient on its own. The type of a segment can also be predicted given the type of its neighbor and their depth difference; this relationship is modeled by $nn2$. For example, a door object is usually ``deeper'' than wall objects. If a segment is a wall and has a higher depth value than its neighbor, then its neighbor has higher probability to be a door object. We also include three MLN potentials. The first suggests that if the angle of a segment is larger than $30^\circ$, then it is likely not a wall object. Similarly, the second formula implies that a segment is of type 'Other' if its angle is larger than $89^\circ$. Finally, $mln3$ encodes the knowledge that neighboring segments are both not likely to be door objects. 

We compare RN-MRFs with Categorical Gaussian MRFs, hybrid MLNs \citep{wang08}, and Neural CRFs \citep{pmlr-v9-do10a}. For hybrid MLNs, we use the HMLN constructed by \citet{wang08}. For CG MRFs and Neural CRFs, we use the grounded graph defined by the above RN-MRF, which could be considered as a CRF model if the length, depth, and angle of all segments are given. To make the comparison between RN-MRFs and these discriminative models fair, we also train our model conditionally and report the result as RN-CRF. Additionally, in order to show the effect of MLN potentials, we include the model both with and without MLN rules.

The dataset consists of 5 distinct laser-scanned maps. For each map, both the endpoint coordinates and relational predicates are provided. In our experiment, we evaluate all models and algorithms using leave-one-out cross validation. All models are trained with PMLE (3000 iterations, 30 sampling points) and tested with EPBP (20 sampling points). The results are shown in Table~\ref{tab:robot_mapping}. 

RN-MRF and RN-CRF perform similarly to Neural CRF, which is likely the result of their somewhat similar reliance on neural networks in the construction of the potential functions. All of the neural network based methods significantly outperform Categorical Gaussian MRFs and the expert-specified hybrid MLN.  One reason for the significant discrepancy is that the local features (length, depth and angle of a segment) are not well-modeled with a unimodal distribution, e.g., in the data set the length distribution of wall objects is closer to bimodal. However, both CG-MRF and HMLN use Gaussians for modeling local continuous features. This suggests that the neural potentials accurately encode this multimodality (\textbf{Q2}). Also note that both RN-MRF and RN-CRF are slightly better than their counterparts without MLN rules. With the neural potentials alone, the RN-MRF appears to capture many of the most important dependencies accurately, but there are also some edge cases, which rarely appear in the dataset, that are difficult to learn. With the addition of a few simple MLN rules, the model better able to account for these dependencies (\textbf{Q3}).

\section{Discussion}
We presented a general relational MRF model that seamlessly handles complex dependencies in relational and hybrid domains and allows for human knowledge to be specified as an inductive bias either in the form of priors or weighted logic rules. When modeling these complex dependencies, RN-MRFs only need to make minimal assumptions on the underlying distribution by utilizing the expressiveness of neural potential function. In addition, we presented a maximum Pseudo-likelihood estimation learning algorithm for general relational MRF/CRF models, that performs well in the context of relational domains and can be trained efficiently. Our empirical evaluations show that the RN-MRF model can be applied to various domains and performs better than non-neural potential based modeling approaches under the same learning conditions. For future work, we plan to explore scaling up this approach to much larger data sets, allowing for richer human knowledge such as preferences and/or qualitative constraints, and exploring the trade-offs/benefits versus neurosymbolic learning methods.

\bibliography{reference}

\end{document}